%% file: 190715_miccai_lesslabels.tex
\documentclass[runningheads]{llncs}
\usepackage{graphicx}
\usepackage{epstopdf}
\usepackage{color, colortbl}
\usepackage[table]{xcolor}
\definecolor{good}{rgb}{0.01, 0.75, 0.24}
	\definecolor{bad}{rgb}{0.55, 0.0, 0.0}
\definecolor{light-gray}{rgb}{0.1, 0.1, 0.1}
\usepackage{array,multirow,graphicx}
\usepackage{booktabs}
\begin{document}
\title{Multi-modal segmentation with missing MR sequences using pre-trained fusion networks}
\titlerunning{Multi-modal segmentation with missing MR sequences}
\author{Karin van Garderen \inst{1,3} \and
Marion Smits\inst{1}  \and
Stefan Klein\inst{1,2}}
\authorrunning{K.A. van Garderen et al.}
\institute{Erasmus MC, Dept. of Radiology and Nuclear Medicine, Rotterdam, the Netherlands \and
Erasmus MC, Dept. of Medical Informatics, Rotterdam, the Netherlands \and
Medical Delta, Delft, the Netherlands}
\maketitle              % typeset the header of the contribution
\begin{abstract}
Missing data is a common problem in machine learning and in retrospective imaging research it is often encountered in the form of missing imaging modalities. We propose to take into account missing modalities in the design and training of neural networks, to ensure that they are capable of providing the best possible prediction even when multiple images are not available. The proposed network combines three modifications to the standard 3D UNet architecture: a training scheme with dropout of modalities, a multi-pathway architecture with fusion layer in the final stage, and the separate pre-training of these pathways. These modifications are evaluated incrementally in terms of performance on full and missing data, using the BraTS multi-modal segmentation challenge. The final model shows significant improvement with respect to the state of the art on missing data and requires less memory during training.
\end{abstract}

\begin{keywords}
convolutional neural network, glioma segmentation, missing data
\end{keywords}

\section{Introduction}
Tumor segmentation is a key task in brain imaging research, as it is a prerequisite for obtaining quantitative features of the tumor. Since manual segmentation by radiologists is time-consuming and prone to inter-observer variation, there is a clear need for effective automatic segmentation methods. Research into these methods for glioma has been accelerated by the recurring BraTS multi-modal segmentation challenge on low-grade glioma (LGG) and glioblastoma (GBM) \cite{6975210}. The best performing methods in recent editions were all based on 3D convolutional neural networks (CNNs) with the encoder-decoder shape of the UNet.

While the BraTS challenge focuses on improving performance, there are practical problems to overcome before automatic segmentation can be applied in practice. One of these challenges is dealing with missing data. The BraTS benchmark contains four MR modalities: a T1-weighted image (T1W), a T1-weighted image with contrast agent (T1WC), a T2-weighted image (T2W) and a T2-weighted FLAIR image (FLAIR), which are co-registered so that corresponding voxels in the image are aligned and a CNN can learn to segment a tumor from the specific combination of modalities. Although these images are complementary, a radiologist is still able to perform a partial segmentation if one of these modalities is missing, while for a CNN this is not guaranteed. Especially in retrospective and multi-center studies it is not unlikely that images are either missing or have quality issues.

%\subsubsection{Related literature}
There are two ways in general to deal with the problem of missing data. The most common way is to impute the missing values by an estimate, which can be as simple as the mean value. More advanced techniques for missing image imputation is to generate a new image from remaining modalities, which can be achieved through neural networks \cite{10.1007/978-3-319-24553-9_83} \cite{JEREZ2010105}. 

However, it is also possible to train a CNN to be inherently robust to missing data. The HeMis model \cite{10.1007/978-3-319-46723-8_54} is an example of this, where the modalities are each passed through a separate pathway before being merged in a so-called abstraction layer which extracts the mean and variance of the resulting features. This network architecture enforces a shared feature representation of the modalities, though it may be of additional value to include a similarity term in the loss function to enforce a true shared representation \cite{10.1007/978-3-319-61188-4_12}.

\subsection{Contribution}
Building on the existing work on shared representations, we provide a careful experimental evaluation of different aspects that make the network robust to missing images. We evaluate four modifications to a state-of-the-art UNet architecture and evaluate their effect incrementally. A first adaptation is to train with missing data in a curriculum learning approach. Secondly, a multi-path architecture is evaluated where the information of different modalities is fused in a later stage. Thirdly, within this architecture, a shared representation layer is compared to a concatenation of feature maps. Finally, we propose a training procedure where each pathway is trained separately before combining them and training the final classification layer. This approach enforces each path to form an informative feature represenation. The separate training also reduces the demand on GPU memory, which is the main bottleneck in state-of-the-art segmentation networks. The modified architectures are compared to the baseline architecture, in a situation where it is trained with the entire dataset but also when it is specifically trained for each combination of modalities.

\section{Methodology}

\begin{figure}[!bt]
 % Caption and label go in the first argument and the figure contents
 % go in the second argument
 \centering
  \includegraphics[width=0.75\linewidth]{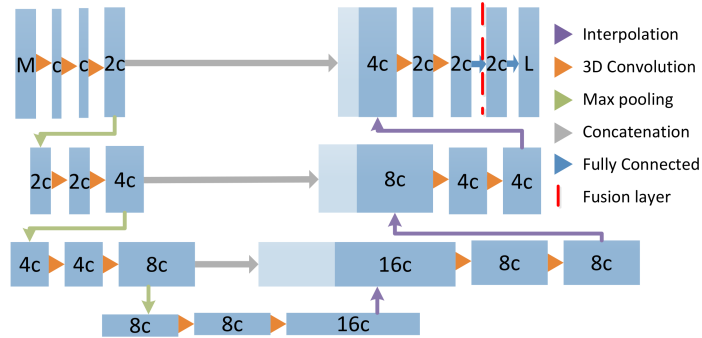}
  
  \caption{Illustration of the UNet architecture. The number of feature maps, as a function of the parameter $c$, is indicated for each step. The fusion and shared representation networks contain one UNet per modality, which are fused at the indicated location. $M$ indicates the number of input modalities and $L$ the number of output labels. In this study $M=4$ and $L=4$.}
  \label{fig:unet}
\end{figure}
\subsection{Network architecture} 
The 3D UNet architecture \cite{10.1007/978-3-319-46723-8_49} is a well-established segmentation network and still was one of the best performing architectures at the most recent 2018 BraTS challenge \cite{2018arXiv180910483I}. Therefore the UNet forms the baseline for our research. One UNet is trained on all modalities and evaluated with missing data, but also a dedicated UNet is trained and evaluated for each specific combination of modalities. The number of trainable parameters in the model depends on the number of feature maps in each convolution, which we chose to parameterize by a single variable $c$. The first convolution has $c$ kernels, and as the size of the feature maps decreases the number of kernels is increased. Fig.~\ref{fig:unet} shows the UNet architecture with the number of feature maps per convolution layer expressed as a multiple of $c$. 

In the reference UNet architecture each 3D convolution block contains a batch normalization, a 3D unpadded convolution layer with kernels of size $3^3$, and Leaky ReLu activation. The last fully connected layers are implemented as a 3D convolution with kernels of size $1^3$. The downsampling step is a max-pooling layer of stride 2 and size $2^3$ and the upsampling is a tri-linear interpolation. For this UNet architecture each target voxel has a receptive field of $88^3$ voxels.

\subsubsection{Modality dropout.}
To make a network robust to missing data it needs to train with missing data. To this end, a specific modality dropout scheme was implemented which removes entire input channels (MR sequences) with a probability $p$. The features from missing sequences are removed by setting the input to zero and scaling the other inputs by $m_o / M$, where $m_o$ is the number of original input images and $M$ is the number of remaining inputs. A curriculum learning approach is used to aid convergence: starting from $p=0.125$ the probability of dropout is doubled every 50 epochs until it reaches $p=0.5$. This method is applied to directly to the input layer in the Dropout network, but also to the fusion layers in the Multipath and SharedRep networks.

\subsubsection{Multipath network.}
In this approach the network has one pathway for each of the $M=4$ modalities and the feature maps of the final convolutional layer are concatenated to an output of $8c$ channels in a fusion layer, which is where the modality dropout is applied. The final prediction is performed again by a $1^3$ convolution layer with $4c$ channels. 

For a fair comparison it is important to consider the number of trainable parameters, which scales quadratically with the number of channels per layer.  To create a multi-path network of the same size as a single reference network, the UNets that form the pathways have half the number of channels per layer. As the UNet was implemented with $c=32$, the separate pathways are a quarter of the size with $c=16$. Note that whereas parameter size scales quadratically, the memory usage scales approximately linear with the number of feature maps. The multi-pathway networks (with $M=4$) therefore require approximately twice the amount of GPU memory during training compared to the single UNet. 

\subsubsection{Shared Representation.} 
The Shared Representation (SharedRep) network is a multi-path network with a specific fusion layer, based on the HeMIS model \cite{10.1007/978-3-319-46723-8_54}. Instead of concatenating, the fusion layer takes the mean and variance of each feature map and therefore encourages a common feature represenation between the modalities. To enable fair comparison to the fusion network, the last layer of each pathway has double the amount of feature maps ($4c$), leading to $8c$ features in the fusion layer. The network is trained with modality dropout of the pathways and the variance is set to zero if only a single pathway is available.

\textbf{Pre-trained paths}
 Pre-training the paths means that a UNet is trained for each individual MR modality and the separate prediction layers are replaced by one fusion layer. These are trained with modality dropout ($p=0.5$), while freezing the parameters of the single pathways. When fusing the pathways with a shared representation layer, the final convolutional layers of the networks are also replaced and trained in order to learn a new shared feature representation. Using the pre-training scheme greatly reduces the demand on GPU memory, as the pathways require a quarter of the memory of the whole network and half that of the full UNet with $c=32$. The combined training scheme took approximately $50\%$ longer than without pre-training, though with parallel training of the paths on separate devices it was even faster than the baseline.

\subsection{Data and preprocessing}
The networks were trained and evaluated on the training set of the BraTS challenge 2018 \cite{bakas2017advancing}, which is a benchmark dataset of pre-operative scans of 278 patients with low-grade glioma (LGG, 75) or glioblastoma (GBM, 203). The images in this benchmark are skull-stripped, co-registered and resampled to a size of 240 by 240 by 155 voxels. The target areas for evaluation are the whole tumor, tumor core and enhancing core. The non-background voxels of each separate image were normalized to zero mean and unit standard deviation. Random patches of $108^3$ voxels were extracted, which correspond to $20^3$ target voxels. With a probability of 50\% a patch was selected from a tumor area, meaning that the center voxel was part of the tumor, and with 50\% probability the center voxel was located outside of the tumor but inside the brain. 

\subsection{Training and evaluation}
The networks were optimized with the Adam optimizer \cite{adam_kingma} and the cross-entropy loss function. An epoch is defined as an iteration over 100 batches with 4 random patches, and the models were trained for 150 epochs. For pre-trained pathways, the separate pathways and the final combination layer were trained for 100 epochs each. The dataset was divided into five cross-validation folds, so that 20\% of the subjects were always selected for testing and never used during training. The folds are random, but the same for each experiment. Evaluation took place on the whole image, although it was classified by the network in patches to limit memory usage. To assess whether the models are indeed more robust to missing data, we evaluated the same models in a situation where any combination of sequences is removed.

\subsection{Visualizing shared representations}
To validate the concept of a shared representation layer in the context of missing data, we would like to know whether the feature representation of such a layer is indeed robust to missing data. We evaluated this in a qualitative way by looking at the t-SNE \cite{maaten2008visualizing} maps of the activations of the final fully connected layer. Feature maps from the final fully connected layer were extracted for 40,000 random voxels originating from 16 random patches. A t-SNE map was computed to map the 64-dimensional feature vectors to a 2D representation. These maps can be interpreted as a representation of the distances between voxels in the specific feature representation of each model. The same set of voxels was used for both maps.

\section{Results}

%\subsection{Quantitative evaluation}
Six networks were trained and evaluated in five-fold cross-validation and, as an additional reference, a dedicated UNet was trained for each combination of sequences. The results are summarized in Table~\ref{tab:results}. On the full dataset, the simple UNet without dropout performs best, and every modification to the network comes with a decreased performance in this case. For missing data scenarios, the regular UNet suffers while the other networks are able to maintain a better performance. None of the networks is able to outperform a dedicated UNet trained for each specific combination of sequences.

There is no architecture that consistently outperforms the others, though the pre-trained multipath networks seem to perform best overall and especially on cases with few available modalities. However, when considering performance on the full dataset, the UNet baseline still performs best and the SharedRep model without pretraining performs better than pretrained paths on the tumor core. Training only with modality dropout greatly decreases performance on the full dataset while only providing minor improvement on missing data.

\begin{table}[!ht]
 % The first argument is the label.
 % The caption goes in the second argument, and the table contents
 % go in the third argument.

  %
  \caption{Numeric results in terms of mean Dice percentage on the three different regions of interest. Color scales are adapted to each region, defined by the best and worst results on that region.}%
  \label{tab:results}
  \centering
  \scriptsize
  \begin{tabular}{r | r | r r r r | r r r r r r | r r r r  }
  
    & \parbox[t]{2mm}{{\rotatebox[origin=l]{90}{\bfseries \scriptsize All }}} & \parbox[t]{2mm}{{\rotatebox[origin=l]{90}{\bfseries \scriptsize All but T1W }}} & \parbox[t]{2mm}{{\rotatebox[origin=l]{90}{\bfseries \scriptsize All but T1WC }}} & \parbox[t]{2mm}{{\rotatebox[origin=l]{90}{\bfseries \scriptsize All but T2W }}} & \parbox[t]{2mm}{{\rotatebox[origin=l]{90}{\bfseries \scriptsize All but FLAIR }}} & \parbox[t]{2mm}{{\rotatebox[origin=l]{90}{\bfseries \scriptsize T2W, FLAIR }}} & \parbox[t]{2mm}{{\rotatebox[origin=l]{90}{\bfseries \scriptsize T1WC, FLAIR }}} & \parbox[t]{2mm}{{\rotatebox[origin=l]{90}{\bfseries \scriptsize T1WC, T2W }}} & \parbox[t]{2mm}{{\rotatebox[origin=l]{90}{\bfseries \scriptsize T1W, FLAIR }}} & \parbox[t]{2mm}{{\rotatebox[origin=l]{90}{\bfseries \scriptsize T1W, T2W }}} & \parbox[t]{2mm}{{\rotatebox[origin=l]{90}{\bfseries \scriptsize T1W, T1WC }}} & \parbox[t]{2mm}{{\rotatebox[origin=l]{90}{\bfseries \scriptsize FLAIR }}} & \parbox[t]{2mm}{{\rotatebox[origin=l]{90}{\bfseries \scriptsize T2W }}} & \parbox[t]{2mm}{{\rotatebox[origin=l]{90}{\bfseries \scriptsize T1WC }}} & \parbox[t]{2mm}{{\rotatebox[origin=l]{90}{\bfseries \scriptsize T1W }}} \\
  \hline
&\multicolumn{15}{r}{\bfseries Whole tumor }\\

\input{results_whole}

\hline
&\multicolumn{15}{r}{\bfseries Tumor core }\\
\input{results_core}
\hline
&\multicolumn{15}{r}{\bfseries Enhancing core }\\
\input{results_enhancing}
 \end{tabular}
\end{table}

\subsection{t-SNE visualizations}
\begin{figure}[!t]
  \centering
  \makebox[\textwidth][c]
  {\includegraphics[width=1.2\textwidth]{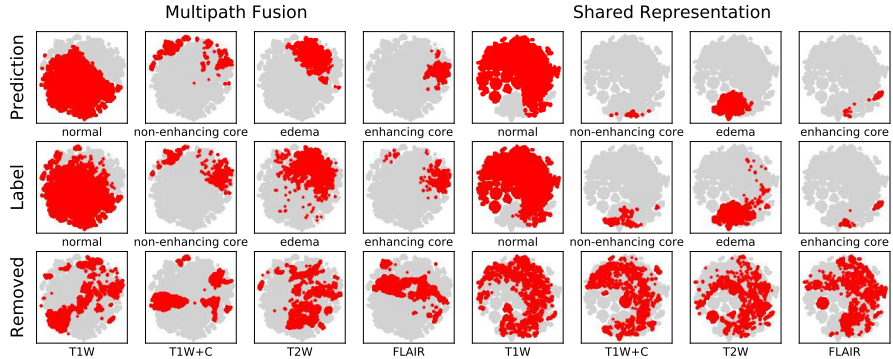}}%
     
  \caption{t-SNE results for pretrained network with fusion by concatenation (left) and shared representation (right). Specific subsets of the voxels are indicated in red. }
   \label{fig:tsne_fusion}
\end{figure}

The resulting t-SNE representations are shown in Fig. \ref{fig:tsne_fusion} for the pretrained Multipath and SharedRep model. The predicted and true labels are highlighted in red, showing that the mapped representation is meaningful to the network prediction and ground truth. Also, the feature maps generated with missing data are highlighted to see whether they lead to distinct feature representations. Whereas the multipath fusion model maps the different missing data scenarios to specific parts of the feature space, the shared representation model seems to have less distinction between complete and incomplete data. This visualization supports the notion that the shared representation layer does indeed lead to a feature representation that is consistent, even when images are removed. 

\section{Discussion and conclusion}
We have carefully evaluated different approaches for training a CNN to be robust to missing imaging modalities, in the context of the BraTs multi-modal segmentation challenge with four MR sequences. Applying modality dropout on the input channels is a simple way to achieve some robustness, but it has a significant impact on performance with full data. More advanced multimodal architectures, with a separate pathway for each modality, give a better balance between performance and robustness. \par
The pathways can be fused either through a simple concatenation or using their statistical moments (mean and variance), thereby enforcing a shared feature representation. Although qualitative visual results show that the shared representation layer forces the feature maps of different combinations of modalities toward a common space, the performance results give no conclusive evidence that it should be preferred over a simple concatenation. The pretraining of the separate paths with a single modality seems to increase the performance mostly in the more difficult cases with fewer modalities. It is also in these cases that a dedicated UNet trained for the specific combination of modalities performs best in comparison, showing that there is still room for improvement. \par
However, it must be noted that the performance achieved by multipath models do not match the best performance on the most recent BraTS training set, as measured on the full dataset. Further improvements on the UNet core are expected to increase the performance further, on both full and partial datasets. \par
The evaluation in this paper has focussed on a systematic comparison of model architectures with the same hyperparameters and size. However, the demand on GPU memory is different between networks. The pre-training of paths in the multipath networks drastically reduces the required memory, so they could be trained with more channels per layer, a larger batch size, a larger patch size or simply a less expensive GPU. It should be preferred for this reason and for its consistent good performance with any combination of modalities.

\section*{Acknowledgements}
%Removed for anonymization.
This work was supported by the Dutch Cancer Society (project number 11026, GLASS-NL), the Dutch Organization for Scientific Research (NWO) and NVIDIA Corporation (by donating a GPU).

\bibliographystyle{splncs04}
\bibliography{midl-samplebibliography}

\end{document}

%% file: results_whole.tex
UNet & \cellcolor{good!48}83 & \cellcolor{good!26}65 & \cellcolor{good!42}78 & \cellcolor{good!37}74 & \cellcolor{bad!1}43 & \cellcolor{good!26}65 & \cellcolor{bad!1}43 & \cellcolor{good!3}46 & \cellcolor{good!23}63 & \cellcolor{bad!26}23 & \cellcolor{bad!32}18 & \cellcolor{bad!8}37 & \cellcolor{bad!18}30 & \cellcolor{bad!38}14 & \cellcolor{bad!50}4 \\
 Dropout & \cellcolor{good!41}77 & \cellcolor{good!41}76 & \cellcolor{good!47}81 & \cellcolor{good!40}76 & \cellcolor{good!19}59 & \cellcolor{good!37}73 & \cellcolor{good!22}62 & \cellcolor{good!19}59 & \cellcolor{good!41}77 & \cellcolor{good!21}61 & \cellcolor{bad!14}33 & \cellcolor{good!9}51 & \cellcolor{good!20}60 & \cellcolor{bad!29}21 & \cellcolor{bad!46}8 \\
 Multipath & \cellcolor{good!48}82 & \cellcolor{good!47}81 & \cellcolor{good!47}82 & \cellcolor{good!42}77 & \cellcolor{good!33}70 & \cellcolor{good!45}80 & \cellcolor{good!38}74 & \cellcolor{good!31}69 & \cellcolor{good!41}77 & \cellcolor{good!32}70 & \cellcolor{bad!3}42 & \cellcolor{good!32}69 & \cellcolor{good!23}63 & \cellcolor{bad!15}32 & \cellcolor{bad!24}25 \\
 SharedRep & \cellcolor{good!49}83 & \cellcolor{good!48}82 & \cellcolor{good!48}82 & \cellcolor{good!43}79 & \cellcolor{good!35}72 & \cellcolor{good!46}81 & \cellcolor{good!37}74 & \cellcolor{good!33}71 & \cellcolor{good!40}76 & \cellcolor{good!33}71 & \cellcolor{good!5}48 & \cellcolor{good!35}72 & \cellcolor{good!31}69 & \cellcolor{bad!10}36 & \cellcolor{bad!19}29 \\
 Multipath + Pretraining & \cellcolor{good!50}84 & \cellcolor{good!49}83 & \cellcolor{good!49}83 & \cellcolor{good!47}82 & \cellcolor{good!39}75 & \cellcolor{good!47}82 & \cellcolor{good!43}78 & \cellcolor{good!37}74 & \cellcolor{good!43}78 & \cellcolor{good!36}73 & \cellcolor{good!15}56 & \cellcolor{good!35}72 & \cellcolor{good!32}70 & \cellcolor{good!6}49 & \cellcolor{bad!0}44 \\
 SharedRep + Pretraining & \cellcolor{good!49}83 & \cellcolor{good!48}83 & \cellcolor{good!48}82 & \cellcolor{good!46}81 & \cellcolor{good!38}74 & \cellcolor{good!46}81 & \cellcolor{good!42}77 & \cellcolor{good!35}72 & \cellcolor{good!43}79 & \cellcolor{good!37}73 & \cellcolor{good!18}58 & \cellcolor{good!39}75 & \cellcolor{good!31}69 & \cellcolor{good!11}52 & \cellcolor{good!0}44 \\
 Dedicated & \cellcolor{good!48}83 & \cellcolor{good!46}81 & \cellcolor{good!47}81 & \cellcolor{good!43}79 & \cellcolor{good!36}73 & \cellcolor{good!44}79 & \cellcolor{good!41}77 & \cellcolor{good!38}74 & \cellcolor{good!39}76 & \cellcolor{good!35}72 & \cellcolor{good!19}59 & \cellcolor{good!37}73 & \cellcolor{good!33}71 & \cellcolor{good!6}49 & \cellcolor{good!5}48 \\

%% file: results_core.tex
 UNet & \cellcolor{good!50}71 & \cellcolor{good!16}47 & \cellcolor{good!10}43 & \cellcolor{good!33}59 & \cellcolor{good!14}46 & \cellcolor{good!9}43 & \cellcolor{bad!16}26 & \cellcolor{bad!0}36 & \cellcolor{bad!3}35 & \cellcolor{bad!19}23 & \cellcolor{bad!15}26 & \cellcolor{bad!12}28 & \cellcolor{bad!13}27 & \cellcolor{bad!40}9 & \cellcolor{bad!50}2 \\
 Dropout & \cellcolor{good!31}57 & \cellcolor{good!33}59 & \cellcolor{bad!5}33 & \cellcolor{good!28}56 & \cellcolor{good!19}50 & \cellcolor{bad!0}36 & \cellcolor{good!8}42 & \cellcolor{good!9}43 & \cellcolor{bad!23}20 & \cellcolor{good!4}39 & \cellcolor{good!9}42 & \cellcolor{bad!12}28 & \cellcolor{good!5}40 & \cellcolor{bad!34}13 & \cellcolor{bad!40}9 \\
 Multipath & \cellcolor{good!47}69 & \cellcolor{good!45}67 & \cellcolor{good!12}44 & \cellcolor{good!40}64 & \cellcolor{good!35}61 & \cellcolor{good!11}44 & \cellcolor{good!32}58 & \cellcolor{good!30}57 & \cellcolor{good!6}40 & \cellcolor{bad!4}33 & \cellcolor{good!14}46 & \cellcolor{bad!1}36 & \cellcolor{bad!4}34 & \cellcolor{bad!7}31 & \cellcolor{bad!16}25 \\
 SharedRep & \cellcolor{good!49}70 & \cellcolor{good!48}69 & \cellcolor{good!10}43 & \cellcolor{good!43}66 & \cellcolor{good!40}64 & \cellcolor{good!7}41 & \cellcolor{good!34}60 & \cellcolor{good!33}59 & \cellcolor{good!3}38 & \cellcolor{good!1}37 & \cellcolor{good!20}50 & \cellcolor{bad!9}30 & \cellcolor{bad!7}31 & \cellcolor{good!3}38 & \cellcolor{bad!23}20 \\
 Multipath + Pretraining & \cellcolor{good!43}66 & \cellcolor{good!42}65 & \cellcolor{good!9}42 & \cellcolor{good!40}64 & \cellcolor{good!40}64 & \cellcolor{good!8}42 & \cellcolor{good!36}61 & \cellcolor{good!35}61 & \cellcolor{good!5}40 & \cellcolor{good!1}37 & \cellcolor{good!25}53 & \cellcolor{bad!4}34 & \cellcolor{bad!1}36 & \cellcolor{good!10}43 & \cellcolor{bad!11}29 \\
 SharedRep + Pretraining & \cellcolor{good!45}67 & \cellcolor{good!44}66 & \cellcolor{good!8}42 & \cellcolor{good!41}64 & \cellcolor{good!39}63 & \cellcolor{good!10}43 & \cellcolor{good!33}59 & \cellcolor{good!34}60 & \cellcolor{good!1}37 & \cellcolor{good!0}37 & \cellcolor{good!25}53 & \cellcolor{bad!4}34 & \cellcolor{bad!11}29 & \cellcolor{good!18}49 & \cellcolor{bad!20}23 \\
 Dedicated & \cellcolor{good!50}71 & \cellcolor{good!40}64 & \cellcolor{good!14}46 & \cellcolor{good!41}64 & \cellcolor{good!38}63 & \cellcolor{good!13}45 & \cellcolor{good!36}61 & \cellcolor{good!39}63 & \cellcolor{good!8}42 & \cellcolor{good!9}43 & \cellcolor{good!28}56 & \cellcolor{good!1}37 & \cellcolor{good!9}43 & \cellcolor{good!10}43 & \cellcolor{bad!16}25 \\

%% file: results_enhancing.tex
UNet & \cellcolor{good!50}63 & \cellcolor{good!14}40 & \cellcolor{bad!42}6 & \cellcolor{good!38}55 & \cellcolor{good!19}43 & \cellcolor{bad!49}2 & \cellcolor{bad!18}21 & \cellcolor{good!6}36 & \cellcolor{bad!42}6 & \cellcolor{bad!46}4 & \cellcolor{bad!11}25 & \cellcolor{bad!40}7 & \cellcolor{bad!42}6 & \cellcolor{bad!42}6 & \cellcolor{bad!47}3 \\
 Dropout & \cellcolor{good!41}57 & \cellcolor{good!39}56 & \cellcolor{bad!40}7 & \cellcolor{good!43}58 & \cellcolor{good!38}55 & \cellcolor{bad!43}5 & \cellcolor{good!11}39 & \cellcolor{good!22}46 & \cellcolor{bad!38}9 & \cellcolor{bad!38}8 & \cellcolor{good!20}44 & \cellcolor{bad!46}4 & \cellcolor{bad!42}6 & \cellcolor{bad!31}13 & \cellcolor{bad!38}9 \\
 Multipath & \cellcolor{good!48}61 & \cellcolor{good!47}61 & \cellcolor{bad!41}7 & \cellcolor{good!43}58 & \cellcolor{good!40}56 & \cellcolor{bad!44}5 & \cellcolor{good!38}55 & \cellcolor{good!35}54 & \cellcolor{bad!40}7 & \cellcolor{bad!39}8 & \cellcolor{good!19}44 & \cellcolor{bad!37}9 & \cellcolor{bad!42}6 & \cellcolor{good!1}33 & \cellcolor{bad!37}9 \\
 SharedRep & \cellcolor{good!48}62 & \cellcolor{good!48}61 & \cellcolor{bad!38}8 & \cellcolor{good!45}60 & \cellcolor{good!42}58 & \cellcolor{bad!40}7 & \cellcolor{good!35}54 & \cellcolor{good!37}55 & \cellcolor{bad!36}10 & \cellcolor{bad!41}7 & \cellcolor{good!26}48 & \cellcolor{bad!44}5 & \cellcolor{bad!44}5 & \cellcolor{good!12}39 & \cellcolor{bad!41}6 \\
 Multipath + Pretraining & \cellcolor{good!48}62 & \cellcolor{good!48}62 & \cellcolor{bad!32}12 & \cellcolor{good!45}60 & \cellcolor{good!46}60 & \cellcolor{bad!32}12 & \cellcolor{good!41}57 & \cellcolor{good!43}58 & \cellcolor{bad!26}16 & \cellcolor{bad!42}6 & \cellcolor{good!29}50 & \cellcolor{bad!25}17 & \cellcolor{bad!50}1 & \cellcolor{good!12}39 & \cellcolor{bad!38}9 \\
 SharedRep + Pretraining & \cellcolor{good!45}60 & \cellcolor{good!46}60 & \cellcolor{bad!35}10 & \cellcolor{good!43}58 & \cellcolor{good!44}59 & \cellcolor{bad!33}12 & \cellcolor{good!36}54 & \cellcolor{good!41}57 & \cellcolor{bad!37}9 & \cellcolor{bad!38}8 & \cellcolor{good!30}50 & \cellcolor{bad!37}9 & \cellcolor{bad!39}8 & \cellcolor{good!26}48 & \cellcolor{bad!37}9 \\
 Dedicated & \cellcolor{good!50}63 & \cellcolor{good!45}60 & \cellcolor{bad!24}17 & \cellcolor{good!50}63 & \cellcolor{good!45}59 & \cellcolor{bad!23}18 & \cellcolor{good!45}60 & \cellcolor{good!42}58 & \cellcolor{bad!24}17 & \cellcolor{bad!29}14 & \cellcolor{good!39}56 & \cellcolor{bad!36}10 & \cellcolor{bad!26}16 & \cellcolor{good!21}45 & \cellcolor{bad!37}9 \\